\documentclass[journal,twoside]{IEEEtran}
\usepackage{cite}
\usepackage{amsmath,amssymb,amsfonts}
\usepackage{algorithmic}
\usepackage{graphicx}
\usepackage{algorithm,algorithmic}
\usepackage{hyperref}
\hypersetup{hidelinks}
\usepackage{textcomp}
\usepackage{mathrsfs}
\usepackage{booktabs}
\usepackage{comment}

\providecommand{\refname}{References}

\usepackage{acronym}
\acrodef{DL}{Deep Learning}
\acrodef{FC}{Functional Connectivity}
\acrodef{MRI}{Magnetic Resonance Imaging}
\acrodef{fMRI}{Functional Magnetic Resonance Imaging}
\acrodef{ASD}{Autism Spectrum Disorder}
\acrodef{ADHD}{Attention-Deficit Hyperactivity Disorder}
\acrodef{ML}{Machine Learning}
\acrodef{FL}{Federated Learning}
\acrodef{XAI}{Explainable AI}
\acrodef{OT}{Optimal Transport}
\acrodef{TKD}{Tucker Decomposition}

\def\BibTeX{{\rm B\kern-.05em{\sc i\kern-.025em b}\kern-.08em
    T\kern-.1667em\lower.7ex\hbox{E}\kern-.125emX}}
\begin{document}
\title{FedDOSE: Federated Learning Framework Decomposing Site Effects for Modeling Brain Dynamic Functional Connectivity}
\author{Deepank Girish, Yi Hao Chan, Yubin Zheng, Sukrit Gupta, and Jagath C. Rajapakse, \IEEEmembership{Fellow, IEEE}
\thanks{This research is supported by AcRF Tier-1 grant RG18/26 of Ministry of Education, Singapore.}
\thanks{Deepank Girish, Yi Hao Chan, and Jagath C. Rajapakse are with the College of Computing and Data Science, Nanyang Technological University, Singapore.}
\thanks{Yubin Zheng is with the School of Computer Science, Shanghai Jiao Tong University, China.}
\thanks{Sukrit Gupta is with the School of Artificial Intelligence and Data Engineering, Indian Institute of Technology Ropar, India.}
\thanks{Corresponding author: Jagath C. Rajapakse (e-mail: ASJagath@ntu.edu.sg).}}

\maketitle

\begin{abstract}

\ac{fMRI} data are often pooled into collaborative multi-site consortia, as deep learning models for analyses require large datasets to generalize well. While \ac{FL} offers a privacy-preserving paradigm for collaborative training, standard approaches continue to struggle with statistical heterogeneity. In particular, site differences pose a key challenge in multi-site data settings. Additionally, existing \ac{FL} approaches for \ac{fMRI} rely on static \ac{FC}, omitting dynamic information in brain networks. To address this, we propose FedDOSE, a novel framework that explicitly decomposes site differences for analysis of dynamic \ac{FC} (DFC). FedDOSE introduces a Modularity-Guided Tucker Decomposition (ModTKD) block to encode high-dimensional DFC tensors and capture modular-level spatio-temporal patterns efficiently. Class-specific prototypes are generated across all sites and subsequently aligned at the global level by using a combination of  barycenter formulation and Procrustes analysis. Extensive experiments for diagnosing \ac{ASD} and \ac{ADHD} on three multi-site resting-state \ac{fMRI}  datasets: ABIDE-I, ABIDE-II, and ADHD-200, demonstrate that FedDOSE achieves competitive performance compared with state-of-the-art methods in \ac{ASD} and \ac{ADHD} detection. Our results highlight its effectiveness in learning robust representations from multi-site fMRI datasets.

\end{abstract}


\section{Introduction}
\label{sec:introduction}
\IEEEPARstart{R}{ecent} advances in deep learning have significantly enhanced fMRI analysis, enabling improved characterization of brain states associated with neurodevelopmental conditions \cite{li2021braingnn, girish2026encoding}. These methods rely on large-scale datasets to achieve better generalizability and accuracy, but such data are often siloed across institutions due to strict privacy regulations \cite{sheller2018multi}. Federated learning (\ac{FL}) enables multiple imaging sites to collaboratively train models while keeping raw data local, thereby mitigating privacy concerns through decentralized training. Recent studies in multi-site \ac{fMRI} analysis have increasingly adopted the \ac{FL} framework \cite{huang2022federated, zhang2024preserving}. However, existing \ac{FL} approaches for \ac{fMRI} analysis face three crucial drawbacks: (1) they primarily focus on static \ac{FC}; (2) they often ignore site-specific characteristics such as phenotypic factors (e.g., age and gender) and scanner variability; (3) they continue to face practical challenges arising from data heterogeneity.  

Static \ac{FC} is computed over an entire scan, lasting several minutes. DFC extends static \ac{FC} by capturing temporal variations in brain connectivity patterns at much faster timescales, on the order of seconds. Although DFC analysis is known to outperform static \ac{FC} in detecting neurodevelopmental disorders \cite{luo2023patterns, zhu2023dynamic}, it also has some drawbacks. DFC estimates are sensitive to methodological choices, such as the window length and stride used in the sliding-window method \cite{zhang2017test}. Estimating DFC over short windows reduces the number of samples available for each estimate, making it susceptible to noise and spurious fluctuations \cite{hindriks2016can}. In addition, DFC typically introduces high dimensionality. If not addressed, this may lead to overfitting and high communication costs in \ac{FL} settings.

Data harmonization (e.g., ComBat \cite{johnson2007adjusting}) and deep generative models (e.g., variational autoencoders \cite{moyer2020scanner}) have been widely used to address heterogeneity across site distributions as they mitigate site-related differences. However, these methods may also remove biologically relevant variance correlated with the site (e.g., age-related connectivity differences across populations). This may induce spurious invariance, which can potentially degrade predictive performance by attenuating informative signals and individual identifiability \cite{wang2023comprehensive}. To better understand site differences, prior work has decomposed them into measurement bias (scanner effects) and sampling bias (participant differences) \cite{yamashita2019harmonization}. However, this approach relies on a small traveling-subject dataset and estimates these biases using simple statistical assumptions. It is imperative to design a principled framework that explicitly separates the underlying sources of site variability. Doing so helps prevent these factors from entangling with disease-relevant signals in \ac{FC} representations.

\begin{figure*}[ht]
\centering
\includegraphics[width=0.9\textwidth]{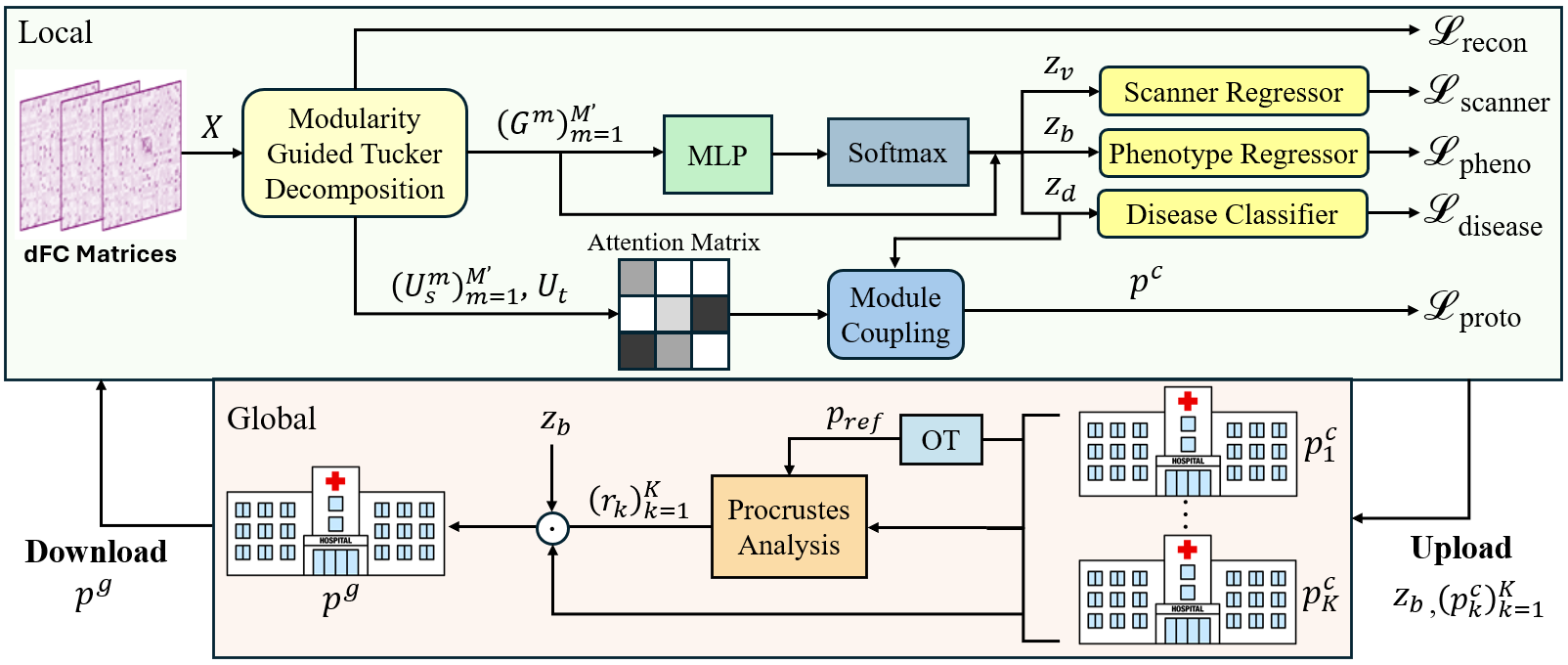}
\caption{Overview of the proposed FedDOSE framework. At the local level, the Modularity-Guided Tucker Decomposition (ModTKD) block extracts compressed modular-level spatiotemporal features from DFC matrices. These features are decomposed into disease, phenotype, and scanner subspaces to separate site differences. Spatial and temporal factors from the ModTKD block are coupled via attention matrix to construct prototypes. At the global level, site prototypes are aligned using OT and Procrustes analysis and then combined with phenotypic saliency scores to form the global prototype.}
\label{fig:overview}
\end{figure*}

To address these challenges, we introduce FedDOSE (Federated Decomposition Of Site Effects), a novel framework that performs a supervised decomposition of site-related variability via DFC. Figure \ref{fig:overview} summarizes the FedDOSE pipeline. Our approach introduces four key innovations: 

\begin{itemize}
    \item We propose the Modularity-Guided \ac{TKD} (ModTKD) block to address the high dimensionality of DFC while also serving as a feature extractor for modular-level spatiotemporal patterns. 
    \item We decompose site differences into disease, phenotypic and scanner subspaces.
    \item We construct class-specific prototypes from the factors obtained via the ModTKD block by capturing spatio-temporal coupling within each module. 
    \item We align all site prototypes at the global level by correcting for distributional heterogeneity using \ac{OT} and Procrustes analysis. After alignment, we construct the global prototype using site-specific phenotypic embeddings.
\end{itemize}

FedDOSE is extensively evaluated on three multi-site rs-\ac{fMRI} datasets covering two neurodevelopmental disorders, namely \ac{ASD} and \ac{ADHD}, comprising a total of 1,368 participants. We show that FedDOSE achieves consistent performance across \ac{ASD} and \ac{ADHD} detection tasks compared with existing \ac{FL} methods, while offering interpretable subspace representations.

\section{Methods}

\subsection{Functional Connectivity Representation}

For each participant, \ac{fMRI} data are represented as a sequence of dynamic \ac{FC} matrices generated using a sliding window over time, indexed by $t \in \{1,2,3,\ldots,T\}$. For each time window, an \ac{FC} matrix $X_t \in \mathbb{R}^{N \times N}$ is constructed, where $N$ denotes the number of brain regions of interest (ROIs). Each entry represents the connectivity strength between a pair of ROIs. $X_t$ is defined as the Pearson correlation of each node with all other nodes. We further construct an adjacency matrix $E_t$ from $X_t$ using an orthogonal minimum spanning tree (OMST) filtering scheme \cite{dimitriadis2017topological}, which removes spurious edges while preserving informative weak connections, yielding a more reliable connectivity representation.

\subsection{Modularity-Guided Tucker Decomposition}

Prior studies have applied \ac{TKD} to \ac{fMRI} datasets \cite{han2021tucker, han2024core}. However, these methods were primarily tailored to static FC-based studies involving small tensor representations. Applying such methods to large tensors like DFC matrices is computationally expensive and may also disrupt the spatial and temporal structure of the data \cite{zhou2026cyclic}. Moreover, tensor decomposition approaches do not incorporate known properties of the connectome, such as the modular structure of \ac{FC} and the presence of brain hubs. Modularity in \ac{FC} refers to the organization of brain networks into groups of densely connected ROIs that have specific brain functions \cite{sporns2016modular}. To address aforementioned challenges, we propose a learnable ModTKD block that captures compressed modular DFC features for downstream analyses.

For each participant, we have a DFC tensor \(X \in \mathbb{R}^{N \times N \times T}\). Since each temporal slice \(X_{t}\) lies on a non-linear space of symmetric positive definite matrices \cite{venkatesh2020comparing}, we first apply the matrix logarithm under Riemannian geometry to map the matrix to the tangent space at the identity matrix. We perform this step because our adopted \ac{TKD} algorithm \cite{tucker1966some} assumes Euclidean structure and applying Euclidean operations on \(X_{t}\) can introduce spurious artifacts \cite{ng2015transport}. The matrix logarithm step is applied before the OMST filtering step, since OMST sparsifies the connectivity matrix, which can alter its eigenvalue spectrum and violate positive definiteness. To reduce computational complexity, we divide the large DFC tensor into smaller modular sub-tensors. For each brain module \(m \in \{1,2,3, \ldots ,M'\}\), we define a binary selection matrix \(B^{m} \in \{0,1\}^{N_m \times N}\), where \(N_m\) denotes the number of ROIs assigned to module \(m\). Using this, we extract the corresponding modular sub-tensor as:
\begin{equation}
    \tilde{X}^{m} = \left(\log(X) \times_1 B^{m}\right) \times_2 B^{m}
\end{equation}
where $\times_n$ denotes tensor multiplication applied along the $n$th dimension of the tensor.

We then apply \ac{TKD} to each sub-tensor corresponding to module \(m\). This step enables faster computation and provides good approximation at a much reduced size. It is given by:
\begin{equation}
    \tilde{X}^{m} \approx G^{m} \times_1 (U_s^{m}) \times_2 (U_s^{m}) \times_3 (U_t)
\end{equation}
where \(U_s^{m} \in \mathbb{R}^{N_m \times R_s}\) is the module-specific spatial factor and \(U_t \in \mathbb{R}^{T \times R_t}\) is the shared temporal factor across all modules.  \(R_s\) and \(R_t\) denote the Tucker ranks controlling the compression. We cyclically apply $U_t$ across all modules during the decomposition step to preserve global temporal dynamics. $G^{m} \in \mathbb{R}^{R_s \times R_s \times R_t}$ denotes the module-specific core tensor, capturing spatial and temporal patterns within module \(m\). The core optimization objective of the ModTKD block is to learn robust feature representations for $G^{m}$. This is achieved by reconstructing the original DFC tensor as follows:  
\begin{equation}
    \hat{X} = \sum_{m=1}^{M'} G^{m} \times_1 ({B^{m}}^\top U_s^{m}) \times_2 ({B^{m}}^\top U_s^{m}) \times_3 (U_t)
\end{equation}

The reconstruction loss \(\mathscr{L}_{\text{recon}}\) is then formulated as:
\begin{equation}
    \mathscr{L}_\text{recon} = \left\|\log(X) - \hat{X} \right\|_F^2 + \sum_{m=1}^{M'}\left\|{U_s^{m}}^\top U_s^{m} - I_{R_{s}} \right\|_F^2
\end{equation}
where \(\|\cdot\|_F\) represents Frobenius norm. The second term in \(\mathscr{L}_\text{recon}\) ensures that each module's spatial factor remains orthonormal during backpropagation. This enforces the spatial patterns to be orthogonal across columns and more interpretable.

Modular core tensors from all modules are aggregated and then passed to the site-effect decomposition block (Section II-C). The residual shared spatial and temporal factors are subsequently used for prototype construction (Section II-D).

\subsection{Decomposition of Site Effects}

In this section, we decompose the modular core tensors into (i) disease-related subspace, (ii) phenotype-preserving subspace, and (iii) scanner-related subspace. We apply softmax-based gating to the aggregated modular core tensors to partition the variance in the core tensors into three distinct subspaces, which is given by: 
\begin{align}
    [M_d, M_b, M_v] &= \mathrm{Softmax}\Big(\text{MLP}\Big(\sum_{m=1}^{M'} G^{m}\Big)\Big), \\
    z_d &= \text{MLP}\Big(\text{vec}\Big(M_d \odot \sum_{m=1}^{M'} G^{m}\Big)\Big)
\end{align}
where \(\odot\) denotes the Hadamard product and \(\text{vec}(\cdot)\) is the vectorization operator. \(z_d\) represents the disease-space embedding obtained by flattening the disease-space mask \(M_d\) and projecting it into a subspace representation. Similar steps are used to obtain the phenotype embedding \(z_b\) and scanner embedding \(z_v\). We then define individual loss terms for each of the three subspaces to perform downstream tasks. 

The disease subspace loss term $\mathscr{L}_{\text{disease}}$ is defined as $\mathscr{L}_{\text{disease}} = \mathscr{L}_{\text{CE}}(h_d(z_d), y)$, where $h_d$ is a prediction head, $y$ denotes the ground-truth disease labels and $\mathscr{L}_{\text{CE}}$ represents the cross-entropy loss. Similarly, the phenotype subspace loss term $\mathscr{L}_{\text{pheno}}$ is defined as $\mathscr{L}_{\text{MSE}}(h_b(z_b), a)$, where $h_b$ is a projection head that maps \(z_b\) to the phenotype space, $a$ denotes the subject-level phenotypic embeddings and $\mathscr{L}_{\text{MSE}}$ represents the mean squared error. In contrast to the other two losses, the scanner subspace loss term $\mathscr{L}_{\text{scanner}}$ is designed to capture scanner-specific information while enforcing invariance to the disease and phenotypic representations. This is performed using a gradient reversal layer \cite{ganin2015unsupervised}, which acts as an identity function during the forward pass and reverses the gradients during backpropagation. It is defined as:
\begin{equation}
\begin{split}
\mathscr{L}_{\text{scanner}} = {}& \mathscr{L}_{\text{MSE}}(h_v(z_v), v) + \mathscr{L}_{\text{MSE}}(h'_v(\mathrm{GRL}(z_d)), v) \\
&+ \mathscr{L}_{\text{MSE}}(h''_v(\mathrm{GRL}(z_b)), v)
\end{split}
\end{equation}
where $v$ denotes the subject-level scanner embeddings. Here, \(h_v\) is a projection head that pushes scanner information into \(z_v\), while \(h'_v\) and \(h''_v\) are adversarial heads on the disease and phenotype branches, applied after the GRL operation. This design choice prevents scanner-related variance from leaking into the disease and phenotypic subspaces. Each source of site variation is encouraged to be captured in each distinct subspace.

\subsection{Spatiotemporal Coupling for Prototype Learning}

Each site computes its class prototypes in the disease subspace using the per-module spatial factors and temporal factor from the ModTKD block. We compute attention logits over all modules to weight each module's contribution to the prototype as: \(\alpha = \mathrm{Softmax}(W' z_d)\)
where \(W'\) is a learnable modular attention matrix and \(\alpha = (\alpha^m)_{m=1}^{M'}\). The disease subspace embedding \(z_d\) is transformed and reshaped into a coupling matrix \(\Sigma\) using an MLP. For each module \(m\), the spatio-temporal coupling map \(F^{m}\) is defined as:
\begin{equation}
    \begin{aligned}
        F^{m} &= U_s^{m} \, \Sigma \, U_{t}^\top 
    \end{aligned}
\end{equation}

\(F^{m}\) captures how the spatial activity of each ROI in module \(m\) evolves over time. For each site \(k\) and subject class \(c\), spatio-temporal prototype \(p_k^c\) is defined as:
\begin{equation}
    p_k^c = \frac{1}{\Omega} \sum_{i=1}^{\Omega} \sum_{m=1}^{M'} \alpha^m_i \, {B^{m}}^\top F^m_i
\end{equation}
where \(\Omega\) denotes the total number of participants belonging to class \(c\) at site \(k\). Since each coupling map \(F^m_i \in \mathbb{R}^{N_m \times T}\) has a module-dependent size, it is projected back to the full ROI space using \({B^{m}}^\top\). This allows the contributions from disjoint modules to be summed, yielding \(p_k^c \in \mathbb{R}^{N \times T}\). The class prototypes are shared across all sites in our framework. Lastly, we introduce a prototype regularization term \(\mathscr{L}_{\text{proto}}\) to use cosine similarity to measure the similarity between local and global prototypes. This encourages each local model to align its prototypes with the global prototype based on directional similarity while being scale-invariant. It is given by:
\begin{equation}
    \mathscr{L}_{\text{proto}} = 1 - \sum_{k=1}^{K} \sum_{c=1}^{C} \frac{\langle p^c_k, p^g \rangle}{\left\|p^c_k\right\|_F \cdot \left\|p^g\right\|_F}
\end{equation}

where \(K\) denotes the total number of sites and \(C\) is the total number of classes. The total local objective $L_{\text{total}}$ at the local level is defined as the combination of five loss functions described above:
\begin{equation}
\begin{split}
    \mathscr{L}_{\text{total}} = {}& \mathscr{L}_{\text{disease}} + \mu\mathscr{L}_{\text{recon}} + \lambda\mathscr{L}_{\text{proto}} \\
    &+ \gamma(\mathscr{L}_{\text{pheno}} + \mathscr{L}_{\text{scanner}}) 
\end{split}
\end{equation} 

We control the contribution of $\mathscr{L}_{\text{recon}}$ using $\mu$ and $\mathscr{L}_{\text{proto}}$ using $\lambda$ as they operate on different scales. We control the combination of $\mathscr{L}_{\text{pheno}}$ and $\mathscr{L}_{\text{scanner}}$ using $\gamma$, as disease detection is our primary task objective and is encouraged to be invariant to site-related variability. After prototype computation, the average phenotypic embedding across all participants is computed. Together with the model parameters and class prototypes from each site, it is transmitted to the global module.

\begin{table*}[t]
\caption{Comparison of classification performance with existing methods on ADHD-200 dataset}
\label{tab:adhd200}
\centering
\small
\setlength{\tabcolsep}{6pt}
\begin{tabular}{lcccccc}
\toprule
\textbf{Models} 
& \multicolumn{2}{c}{\textbf{NYU}} 
& \multicolumn{2}{c}{\textbf{OHSU}} 
& \multicolumn{2}{c}{\textbf{PKU}} \\
\cmidrule(lr){2-3} \cmidrule(lr){4-5} \cmidrule(lr){6-7}
& \shortstack{Site Acc} & \shortstack{Global Acc}
& \shortstack{Site Acc} & \shortstack{Global Acc}
& \shortstack{Site Acc} & \shortstack{Global Acc} \\
\midrule
Local 
& 0.58 \(\pm\) 0.07 & N/A 
& 0.60 \(\pm\) 0.05 & N/A 
& 0.62 \(\pm\) 0.06 & N/A \\
Centralised 
& N/A & 0.62 \(\pm\) 0.05 
& N/A & 0.64 \(\pm\) 0.06 
& N/A & 0.66 \(\pm\) 0.06 \\
\midrule
FedAvg \cite{mcmahan2017communication}
& 0.54 \(\pm\) 0.06*\textasciicircum & 0.51 \(\pm\) 0.05*\textasciicircum 
& 0.55 \(\pm\) 0.08\,\;\textasciicircum & 0.52 \(\pm\) 0.06*\textasciicircum 
& 0.53 \(\pm\) 0.08*\textasciicircum & 0.50 \(\pm\) 0.07*\textasciicircum \\
FedProto \cite{tan2022fedproto}
& 0.56 \(\pm\) 0.08\,\;\textasciicircum & N/A 
& 0.56 \(\pm\) 0.10\,\;\textasciicircum & N/A 
& 0.55 \(\pm\) 0.09*\textasciicircum & N/A \\
TDPFed \cite{wang2022tensor}
& 0.57 \(\pm\) 0.07\,\;\textasciicircum & 0.53 \(\pm\) 0.06\,\;\textasciicircum
& 0.58 \(\pm\) 0.09\,\;\textasciicircum & 0.54 \(\pm\) 0.08\,\;\textasciicircum
& 0.56 \(\pm\) 0.08*\textasciicircum & 0.53 \(\pm\) 0.07*\textasciicircum \\
GAFD \cite{wang2025graph}
& 0.57 \(\pm\) 0.09\,\;\textasciicircum & 0.54 \(\pm\) 0.06\,\;\textasciicircum 
& 0.59 \(\pm\) 0.07\,\;\textasciicircum & 0.55 \(\pm\) 0.08\,\;\textasciicircum 
& 0.58 \(\pm\) 0.07*\textasciicircum & 0.56 \(\pm\) 0.07*\textasciicircum \\
FedGST \cite{mao2023fedgst}
& 0.59 \(\pm\) 0.08\,\;\phantom{\textasciicircum} & 0.55 \(\pm\) 0.06\,\;\textasciicircum 
& 0.60 \(\pm\) 0.09\,\;\phantom{\textasciicircum} & 0.57 \(\pm\) 0.06\,\;\textasciicircum 
& 0.61 \(\pm\) 0.06\,\;\textasciicircum & 0.58 \(\pm\) 0.04*\textasciicircum \\
FedAli \cite{ek2024fedali}
& 0.60 \(\pm\) 0.08\,\;\phantom{\textasciicircum} & 0.57 \(\pm\) 0.04\,\;\phantom{\textasciicircum} 
& 0.61 \(\pm\) 0.06\,\;\phantom{\textasciicircum}  & 0.58 \(\pm\) 0.07\,\;\phantom{\textasciicircum} 
& 0.63 \(\pm\) 0.05\,\;\textasciicircum & 0.60 \(\pm\) 0.06\,\;\textasciicircum \\
FedGMKD \cite{zhang2024fedgmkd} 
& 0.61 \(\pm\) 0.07\,\;\phantom{\textasciicircum} & 0.57 \(\pm\) 0.05\,\;\phantom{\textasciicircum} 
& 0.62 \(\pm\) 0.06\,\;\phantom{\textasciicircum}  & 0.59 \(\pm\) 0.07\,\;\phantom{\textasciicircum} 
& 0.64 \(\pm\) 0.05\,\;\phantom{\textasciicircum} & 0.61 \(\pm\) 0.04\,\;\textasciicircum \\
FS\(^2\)G \cite{mao2024toward}
& 0.61 \(\pm\) 0.06\,\;\phantom{\textasciicircum} & 0.58 \(\pm\) 0.06\,\;\phantom{\textasciicircum} 
& 0.63 \(\pm\) 0.05\,\;\phantom{\textasciicircum} & 0.59 \(\pm\) 0.06\,\;\phantom{\textasciicircum} 
& 0.65 \(\pm\) 0.07\,\;\phantom{\textasciicircum} & 0.62 \(\pm\) 0.06\,\;\phantom{\textasciicircum} \\
FedDOSE
& \textbf{0.64 \(\pm\) 0.07\,\;\phantom{\textasciicircum}} & \textbf{0.60 \(\pm\) 0.06\,\;\phantom{\textasciicircum}} 
& \textbf{0.65 \(\pm\) 0.07\,\;\phantom{\textasciicircum}} & \textbf{0.62 \(\pm\) 0.06\,\;\phantom{\textasciicircum}} 
& \textbf{0.68 \(\pm\) 0.06\,\;\phantom{\textasciicircum}} & \textbf{0.65 \(\pm\) 0.05\,\;\phantom{\textasciicircum}} \\
\bottomrule
\end{tabular}
\vspace{8pt}

\caption{Comparison of classification performance with existing methods on ABIDE-I dataset}
\label{tab:abide1}
\centering
\small
\setlength{\tabcolsep}{6pt}
\begin{tabular}{lcccccc}
\toprule
\textbf{Models} 
& \multicolumn{2}{c}{\textbf{NYU}} 
& \multicolumn{2}{c}{\textbf{UCLA}} 
& \multicolumn{2}{c}{\textbf{UM}} \\
\cmidrule(lr){2-3} \cmidrule(lr){4-5} \cmidrule(lr){6-7}
& \shortstack{Site Acc} & \shortstack{Global Acc}
& \shortstack{Site Acc} & \shortstack{Global Acc}
& \shortstack{Site Acc} & \shortstack{Global Acc} \\
\midrule
Local 
& 0.64 \(\pm\) 0.05 & N/A 
& 0.68 \(\pm\) 0.06 & N/A 
& 0.70 \(\pm\) 0.05 & N/A \\
Centralised 
& N/A & 0.69 \(\pm\) 0.06 
& N/A & 0.74 \(\pm\) 0.05 
& N/A & 0.75 \(\pm\) 0.04 \\
\midrule
FedAvg \cite{mcmahan2017communication}
& 0.58 \(\pm\) 0.08*\textasciicircum & 0.54 \(\pm\) 0.08*\textasciicircum 
& 0.65 \(\pm\) 0.08\,\;\textasciicircum & 0.60 \(\pm\) 0.07*\textasciicircum 
& 0.62 \(\pm\) 0.08*\textasciicircum & 0.57 \(\pm\) 0.07*\textasciicircum \\
FedProto \cite{tan2022fedproto}
& 0.59 \(\pm\) 0.07*\textasciicircum & N/A 
& 0.63 \(\pm\) 0.08*\textasciicircum & N/A 
& 0.64 \(\pm\) 0.07*\textasciicircum & N/A \\
TDPFed \cite{wang2022tensor}
& 0.60 \(\pm\) 0.06*\textasciicircum & 0.56 \(\pm\) 0.08*\textasciicircum
& 0.66 \(\pm\) 0.06*\textasciicircum & 0.62 \(\pm\) 0.07*\textasciicircum 
& 0.67 \(\pm\) 0.07\,\;\textasciicircum & 0.62 \(\pm\) 0.06*\textasciicircum \\
GAFD \cite{wang2025graph}
& 0.62 \(\pm\) 0.06*\textasciicircum & 0.59 \(\pm\) 0.05*\textasciicircum 
& 0.69 \(\pm\) 0.08\,\;\textasciicircum & 0.64 \(\pm\) 0.07\,\;\textasciicircum 
& 0.70 \(\pm\) 0.07\,\;\textasciicircum & 0.66 \(\pm\) 0.06\,\;\textasciicircum \\
FedGST \cite{mao2023fedgst}
& 0.65 \(\pm\) 0.06\,\;\textasciicircum & 0.62 \(\pm\) 0.06\,\;\textasciicircum 
& 0.71 \(\pm\) 0.07\,\;\phantom{\textasciicircum} & 0.65 \(\pm\) 0.06\,\;\textasciicircum 
& 0.71 \(\pm\) 0.06\,\;\textasciicircum & 0.68 \(\pm\) 0.06\,\;\textasciicircum \\
FedAli \cite{ek2024fedali}
& 0.66 \(\pm\) 0.05\,\;\textasciicircum & 0.63 \(\pm\) 0.06\,\;\phantom{\textasciicircum} 
& 0.72 \(\pm\) 0.06\,\;\phantom{\textasciicircum} & 0.67 \(\pm\) 0.06\,\;\phantom{\textasciicircum} 
& 0.73 \(\pm\) 0.06\,\;\phantom{\textasciicircum} & 0.69 \(\pm\) 0.07\,\;\phantom{\textasciicircum} \\
FedGMKD \cite{zhang2024fedgmkd}
& 0.68 \(\pm\) 0.06\,\;\phantom{\textasciicircum} & 0.64 \(\pm\) 0.05\,\;\phantom{\textasciicircum} 
& 0.73 \(\pm\) 0.06\,\;\phantom{\textasciicircum} & 0.68 \(\pm\) 0.06\,\;\phantom{\textasciicircum} 
& 0.73 \(\pm\) 0.07\,\;\phantom{\textasciicircum} & 0.70 \(\pm\) 0.06\,\;\phantom{\textasciicircum} \\
FS\(^2\)G \cite{mao2024toward}
& 0.69 \(\pm\) 0.05\,\;\phantom{\textasciicircum} & 0.65 \(\pm\) 0.04\,\;\phantom{\textasciicircum} 
& \textbf{0.76 \(\pm\) 0.05\,\;\phantom{\textasciicircum}} & 0.70 \(\pm\) 0.05\,\;\phantom{\textasciicircum} 
& 0.76 \(\pm\) 0.06\,\;\phantom{\textasciicircum} & 0.72 \(\pm\) 0.05\,\;\phantom{\textasciicircum} \\
FedDOSE 
& \textbf{0.71 \(\pm\) 0.06\,\;\phantom{\textasciicircum}} & \textbf{0.67 \(\pm\) 0.05\,\;\phantom{\textasciicircum}} 
& 0.75 \(\pm\) 0.06\,\;\phantom{\textasciicircum} & \textbf{0.71 \(\pm\) 0.05\,\;\phantom{\textasciicircum}} 
& \textbf{0.77 \(\pm\) 0.07\,\;\phantom{\textasciicircum}} & \textbf{0.73 \(\pm\) 0.06\,\;\phantom{\textasciicircum}} \\
\bottomrule
\end{tabular}
\end{table*}

\subsection{Prototype Alignment via OT}
Upon receiving the class prototypes, model parameters and phenotypic embedding, the global module performs a two-stage aggregation. In the first stage, it aggregates the local model parameters via standard federated averaging \cite{mcmahan2017communication}, yielding the updated global model: 
\(f_g = \frac{1}{K} \sum_{k=1}^{K} f_k\)

In the second stage, aggregating prototypes using simple averaging is suboptimal. This is because Tucker decompositions are performed independently across sites, which leads to misalignment in the ordering of latent components. This introduces heterogeneity among prototypes. For instance, a spatial factor corresponding to the default mode network (DMN) in one site may correspond to a different functional network, such as the salience network (SN), in another site. To address this, we first align local prototypes using generalized Procrustes analysis \cite{gower1975generalized}, where \ac{OT} is used to infer cross-site mappings and define a shared reference space. Procrustes-based methods have been widely used in neuroscience, particularly in \ac{fMRI} preprocessing, where they have been used to align functional and anatomical brain structures \cite{andreella2023procrustes}. As it is reasonable to assume that local prototypes are drawn from different distributions, we employ an \ac{OT} barycenter formulation \cite{kolesov2024energy} as it provides a principled approach for averaging the distributions of prototypes across sites:
\begin{equation}
p_{\mathrm{ref}} = \arg\min_{p} \sum_{k=1}^K \texttt{W}_2^2(p, p_k)
\end{equation}
where \(\texttt{W}_2^2\) denotes the squared 2-Wasserstein distance function and \(p\) denotes the candidate reference prototypes. \(p_{\mathrm{ref}}\) denotes the reference distribution that minimizes the aggregate Wasserstein distance across all site prototype distributions. Once the reference matrix is obtained, each site's prototype is aligned to it using the Procrustes problem. It is formulated as:
\begin{equation}
r_k = \arg\min_{r} \|p_{\mathrm{ref}} - p_k r\|_F
\end{equation}

where \(r_k\) is the rotational matrix for site \(k\)'s prototype. When multiplied by the site prototype, it aligns it as closely as possible to the reference prototype. \(r\) denotes the set of all possible rotation matrices and is the variable over which optimization is performed. The closed-form solution is obtained via singular value decomposition (SVD) of the cross-covariance matrix \(p_k^\top p_{\mathrm{ref}}\). The optimal rotation is then given by \(r_k = U'_k V_k^{'\top}\), where \(U'_k\) and \(V'_k\) are the left and right singular vectors obtained from the SVD. Finally, we can perform global prototype aggregation, defined as:
\begin{equation}
p^g = \sum_{k=1}^K w_k \cdot p_k r_k
\end{equation}

By aligning all site prototypes to a common reference coordinate system, we obtain geometrically and biologically meaningful representations. Here, \(w_k\) denotes the phenotype-based saliency score for site \(k\). It determines the contribution of each site to the global prototype. It is computed as the normalized inverse mean distance of its averaged phenotypic embedding to those of other sites. Finally, the global prototype \(p_g\) is distributed back to all the sites for next round of end-to-end federated training.

\section{Results}
\label{sec:experiments}

\subsection{Datasets and preprocessing}

We evaluated our framework on three public multi-site rs-\ac{fMRI} datasets, demonstrating its generalizability:

\noindent \textbf{ABIDE-I \cite{di2014autism}:} Autism Brain Imaging Data Exchange I (ABIDE-I) dataset contains 387 rs-\ac{fMRI} scans from individuals diagnosed with \ac{ASD} and 436 typically developing controls, collected from 20 sites. After quality control and site selection, the final cohort comprises 146 individuals with \ac{ASD} and 177 typically developing controls from three sites: NYU (88 NC / 73 ASD), UCLA (39 NC / 36 ASD), and UM (50 NC / 37 ASD). These sites were chosen as they have been used to benchmark model performance in other \ac{FL} studies on rs-\ac{fMRI} data.
 
\noindent \textbf{ABIDE-II \cite{di2017enhancing}:} Similar to ABIDE-I, the Autism Brain Imaging Data Exchange II (ABIDE-II) dataset comprises 107 individuals with \ac{ASD} and 111 typically developing controls from three sites: BNI (29 NC / 29 ASD), EMC (27 NC / 27 ASD), and GU (55 NC / 51 ASD).

\noindent \textbf{ADHD-200 \cite{brown2012adhd}:} ADHD-200 dataset was used to further validate our findings. After quality control and site selection, the final cohort comprises 374 individuals with \ac{ADHD} and 453 typically developing controls. We selected NYU (212 NC / 184 ADHD), OHSU (125 NC / 112 ADHD), and PKU (116 NC / 78 ADHD), as the remaining sites suffered from low sample size (NI) and severe class imbalance (KKI, WUSTL).

We truncated all sites to the minimum number of time points within each dataset, resulting in a common $T$ for that dataset. Following recommended practices \cite{leonardi2015spurious, zhang2017test}, we constructed DFC matrices by segmenting BOLD signals into overlapping windows of 60 timepoints with a stride of 1 timepoint, yielding 61 windows per participant for ABIDE-I and ABIDE-II and 19 windows per participant for ADHD-200. Power atlas \cite{power2011functional} was used to define 264 ROIs. We used eight modules from the Power atlas for the ModTKD block. FC matrices were computed using Pearson correlation between the mean time series of each pair of ROIs. Resulting DFC matrices from each sliding window were used as model inputs. From each dataset, the phenotypic embedding is constructed by concatenating the participant's age with an encoded representation of gender. Similarly, the scanner embedding is constructed by concatenating encoded site labels, encoded scanner-type information, and subject-specific temporal resolution.

\subsection{Baselines}

We benchmarked FedDOSE against state-of-the-art \ac{FL} models from both fMRI-specific and generic (non-fMRI) domains. These models include FedAvg \cite{mcmahan2017communication}, FedProto, TDPFed \cite{wang2022tensor}, GAFD \cite{wang2025graph}, FedGST \cite{mao2023fedgst}, FedAli \cite{ek2024fedali}, FedGMKD \cite{zhang2024fedgmkd}, and FS\(^2\)G \cite{mao2024toward}. FedAvg is the standard federated optimization method that constructs a global model by performing weighted parameter averaging across clients. In FedProto, class-level feature prototypes are exchanged instead of full model parameters. TDPFed uses \ac{TKD} to design a tensorised local model and decouples personalized model optimisation from global model learning. GAFD introduced graph augmentation for static \ac{FC} brain networks and applied knowledge distillation to improve multi-site aggregation. FedGST extracts spatiotemporal patterns using a GCN and DFC, with the temporal model trained on the client side and the spatial model trained on the server side. FedAli utilizes \ac{OT} to enforce consistency between local and global representations, reducing client drift and improving convergence stability. FedGMKD combines a knowledge distillation with differential aggregation to enable a personalized prototype-based \ac{FL} framework. Lastly, FS\(^2\)G \cite{mao2024toward} employs UniteGCN to extract spatial features for disease prediction and uses attention to fuse the outputs of multiple GCNs. 

\begin{table*}[t]
\caption{Comparison of classification performance with existing methods on ABIDE-II dataset}
\label{tab:abide2}
\centering
\small
\setlength{\tabcolsep}{6pt}
\begin{tabular}{lcccccc}
\toprule
\textbf{Models} 
& \multicolumn{2}{c}{\textbf{BNI}} 
& \multicolumn{2}{c}{\textbf{EMC}} 
& \multicolumn{2}{c}{\textbf{GU}} \\
\cmidrule(lr){2-3} \cmidrule(lr){4-5} \cmidrule(lr){6-7}
& \shortstack{Site Acc} & \shortstack{Global Acc}
& \shortstack{Site Acc} & \shortstack{Global Acc}
& \shortstack{Site Acc} & \shortstack{Global Acc} \\
\midrule
Local 
& 0.63 \(\pm\) 0.05 & N/A 
& 0.59 \(\pm\) 0.04 & N/A 
& 0.62 \(\pm\) 0.05 & N/A \\
Centralised 
& N/A &  0.69 \(\pm\) 0.04  
& N/A &  0.64 \(\pm\) 0.04  
& N/A &  0.67 \(\pm\) 0.06 \\
\midrule
FedAvg \cite{mcmahan2017communication}
& 0.53 \(\pm\) 0.09*\textasciicircum & 0.51 \(\pm\) 0.08*\textasciicircum 
& 0.54 \(\pm\) 0.11*\textasciicircum & 0.51 \(\pm\) 0.09*\textasciicircum 
& 0.54 \(\pm\) 0.09*\textasciicircum & 0.52 \(\pm\) 0.08*\textasciicircum \\
FedProto \cite{tan2022fedproto}
& 0.56 \(\pm\) 0.08*\textasciicircum & N/A 
& 0.54 \(\pm\) 0.09*\textasciicircum & N/A 
& 0.56 \(\pm\) 0.09*\textasciicircum & N/A \\
TDPFed \cite{wang2022tensor}
& 0.58 \(\pm\) 0.09*\textasciicircum & 0.53 \(\pm\) 0.08*\textasciicircum  
& 0.56 \(\pm\) 0.09*\textasciicircum & 0.52 \(\pm\) 0.08*\textasciicircum 
& 0.57 \(\pm\) 0.07*\textasciicircum & 0.53 \(\pm\) 0.07*\textasciicircum \\
GAFD \cite{wang2025graph}
& 0.59 \(\pm\) 0.07*\textasciicircum & 0.57 \(\pm\) 0.08\,\;\textasciicircum 
& 0.58 \(\pm\) 0.08*\textasciicircum & 0.56 \(\pm\) 0.09\,\;\textasciicircum 
& 0.59 \(\pm\) 0.06*\textasciicircum & 0.56 \(\pm\) 0.07\,\;\textasciicircum \\
FedGST \cite{mao2023fedgst}
& 0.62 \(\pm\) 0.08\,\;\textasciicircum & 0.60 \(\pm\) 0.07\,\;\textasciicircum 
& 0.60 \(\pm\) 0.08\,\;\textasciicircum & 0.57 \(\pm\) 0.08\,\;\textasciicircum 
& 0.61 \(\pm\) 0.06\,\;\textasciicircum & 0.58 \(\pm\) 0.06\,\;\textasciicircum \\
FedAli \cite{ek2024fedali}
& 0.65 \(\pm\) 0.08\,\;\phantom{\textasciicircum} & 0.61 \(\pm\) 0.06\,\;\textasciicircum 
& 0.63 \(\pm\) 0.08\,\;\textasciicircum & 0.59 \(\pm\) 0.06\,\;\phantom{\textasciicircum} 
& 0.62 \(\pm\) 0.07\,\;\textasciicircum & 0.60 \(\pm\) 0.05\,\;\textasciicircum \\
FedGMKD \cite{zhang2024fedgmkd}
& 0.67 \(\pm\) 0.07\,\;\phantom{\textasciicircum} & 0.63 \(\pm\) 0.06\,\;\phantom{\textasciicircum} 
& 0.65 \(\pm\) 0.06\,\;\phantom{\textasciicircum} & 0.61 \(\pm\) 0.07\,\;\phantom{\textasciicircum} 
& 0.64 \(\pm\) 0.06\,\;\phantom{\textasciicircum} & 0.60 \(\pm\) 0.05\,\;\textasciicircum \\
FS\(^2\)G \cite{mao2024toward}
& 0.66 \(\pm\) 0.07\,\;\phantom{\textasciicircum} & 0.64 \(\pm\) 0.07\,\;\phantom{\textasciicircum} 
& 0.64 \(\pm\) 0.07\,\;\phantom{\textasciicircum} & 0.61 \(\pm\) 0.05\,\;\phantom{\textasciicircum} 
& 0.65 \(\pm\) 0.06\,\;\phantom{\textasciicircum} & 0.59 \(\pm\) 0.06\,\;\textasciicircum \\
FedDOSE 
& \textbf{0.70 \(\pm\) 0.05\,\;\phantom{\textasciicircum}} & \textbf{0.67 \(\pm\) 0.06\,\;\phantom{\textasciicircum}} 
& \textbf{0.69 \(\pm\) 0.06\,\;\phantom{\textasciicircum}} & \textbf{0.63 \(\pm\) 0.05\,\;\phantom{\textasciicircum}} 
& \textbf{0.68 \(\pm\) 0.05\,\;\phantom{\textasciicircum}} & \textbf{0.64 \(\pm\) 0.05\,\;\phantom{\textasciicircum}} \\
\bottomrule
\end{tabular}

\vspace{8pt}

\caption{Ablation study of FedDOSE on ABIDE-I dataset: ASD prediction accuracies}
\label{tab:ablation}
\centering
\small
\setlength{\tabcolsep}{6pt}
\begin{tabular}{lcccccc}
\toprule
\textbf{Models} 
& \multicolumn{2}{c}{\textbf{NYU}} 
& \multicolumn{2}{c}{\textbf{UCLA}} 
& \multicolumn{2}{c}{\textbf{UM}} \\
\cmidrule(lr){2-3} \cmidrule(lr){4-5} \cmidrule(lr){6-7}
& \shortstack{Site Acc} & \shortstack{Global Acc}
& \shortstack{Site Acc} & \shortstack{Global Acc}
& \shortstack{Site Acc} & \shortstack{Global Acc} \\
\midrule
w/o \(\mathscr{L}_{\text{pheno}}, \mathscr{L}_{\text{scanner}}\)
& 0.60 \(\pm\) 0.06*\textasciicircum & 0.55 \(\pm\) 0.05*\textasciicircum 
& 0.61 \(\pm\) 0.06*\textasciicircum & 0.53 \(\pm\) 0.06*\textasciicircum 
& 0.61 \(\pm\) 0.06*\textasciicircum & 0.55 \(\pm\) 0.05*\textasciicircum \\
w/o \(\mathscr{L}_{\text{recon}}\) 
& 0.62 \(\pm\) 0.06*\textasciicircum & 0.57 \(\pm\) 0.04*\textasciicircum 
& 0.63 \(\pm\) 0.07*\textasciicircum & 0.56 \(\pm\) 0.05*\textasciicircum 
& 0.64 \(\pm\) 0.06*\textasciicircum & 0.56 \(\pm\) 0.04*\textasciicircum \\
w/o Procrustes 
& 0.65 \(\pm\) 0.05\,\;\textasciicircum & 0.59 \(\pm\) 0.04*\textasciicircum 
& 0.67 \(\pm\) 0.05*\textasciicircum & 0.60 \(\pm\) 0.04*\textasciicircum  
& 0.68 \(\pm\) 0.05*\textasciicircum & 0.61 \(\pm\) 0.06*\textasciicircum \\
w/o \(\mathscr{L}_{\text{pheno}}\)
& 0.65 \(\pm\) 0.06\,\;\textasciicircum & 0.60 \(\pm\) 0.05\,\;\textasciicircum 
& 0.66 \(\pm\) 0.06*\textasciicircum & 0.58 \(\pm\) 0.06*\textasciicircum 
& 0.69 \(\pm\) 0.07\,\;\textasciicircum & 0.62 \(\pm\) 0.05*\textasciicircum \\
w/o OT
& 0.67 \(\pm\) 0.07\,\;\phantom{\textasciicircum} & 0.61 \(\pm\) 0.05\,\;\textasciicircum
& 0.70 \(\pm\) 0.07\,\;\phantom{\textasciicircum} & 0.64 \(\pm\) 0.06\,\;\textasciicircum 
& 0.72 \(\pm\) 0.07\,\;\phantom{\textasciicircum} & 0.65 \(\pm\) 0.05*\textasciicircum \\
w/o \(\mathscr{L}_{\text{proto}}\) 
& 0.69 \(\pm\) 0.07\,\;\phantom{\textasciicircum}  & 0.64 \(\pm\) 0.06\,\;\phantom{\textasciicircum} 
& 0.71 \(\pm\) 0.06\,\;\phantom{\textasciicircum} & 0.66 \(\pm\) 0.07\,\;\textasciicircum 
& 0.74 \(\pm\) 0.08\,\;\phantom{\textasciicircum} & 0.68 \(\pm\) 0.07\,\;\phantom{\textasciicircum} \\
FedDOSE 
& \textbf{0.71 \(\pm\) 0.06\,\;\phantom{\textasciicircum}} & \textbf{0.67 \(\pm\) 0.05\,\;\phantom{\textasciicircum}} 
& \textbf{0.75 \(\pm\) 0.06\,\;\phantom{\textasciicircum}} & \textbf{0.71 \(\pm\) 0.05\,\;\phantom{\textasciicircum}} 
& \textbf{0.77 \(\pm\) 0.07\,\;\phantom{\textasciicircum}} & \textbf{0.73 \(\pm\) 0.06\,\;\phantom{\textasciicircum}} \\
\bottomrule
\end{tabular}
\end{table*}

\subsection{Implementation details}

Each experiment was repeated with five random seeds, and every dataset was split 60/20/20 into training, validation and test sets. This split was fixed across all methods and seeds, and participants with repeated scans or sessions were assigned to a single split to prevent data leakage between training and testing. Models were trained using Adam optimizer with a learning rate of 0.005. The federated process ran for 40 rounds with 5 local epochs per round. Tuning was performed using grid search, where the tuned hyperparameters include $\mu \in \{0.01, 0.05, 0.1\}$, $\gamma \in \{0.1, 0.3, 0.5\}$, and $\lambda \in \{0.1, 0.2, 0.3\}$. Hyperparameters were selected on the validation split and all reported numbers are test-set results. We used a batch size of 8. All models were implemented in PyTorch 2.1 (Python 3.10) and trained on a single NVIDIA A100 GPU. We also compare FedDOSE against two standard learning approaches: Local and Centralised. In the Local setting, sites train independently on their own datasets. In the Centralised setting, a global model is trained on pooled data from all sites. The other baseline models were trained using the same hyperparameter tuning budget. We evaluate performance using two metrics: site accuracy (Site Acc) and global accuracy (Global Acc). Site accuracy is the classification accuracy of each site's locally adapted model on its own data after federated training. Global accuracy corresponds to the accuracy of the global federated model evaluated separately on each site's test set.

\subsection{Performance comparison with the baselines}

\begin{figure*}[t]
\centering
\includegraphics[width=0.85\textwidth]{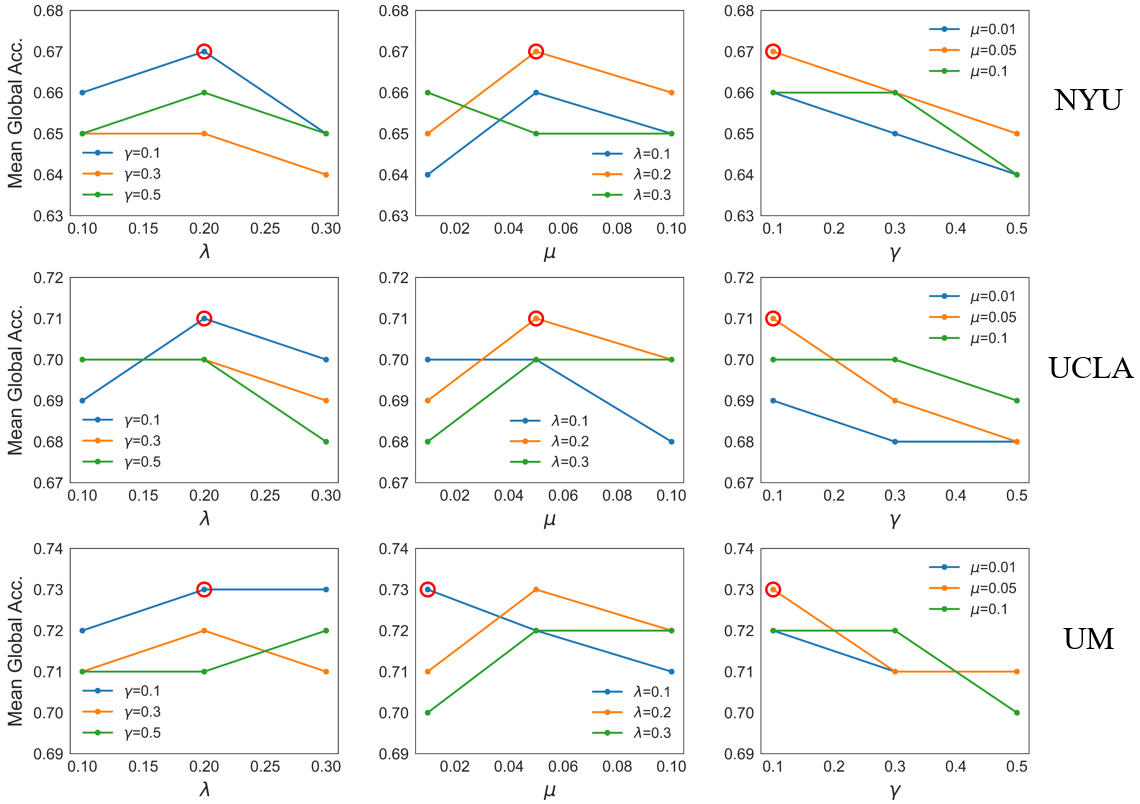}
\caption{FedDOSE mean global accuracy on the ABIDE-I dataset for different hyperparameter pairs: $(\lambda, \gamma)$ with $\mu$ fixed (Left), $(\mu, \lambda)$ with $\gamma$ fixed (Middle), and $(\gamma, \mu)$ with $\lambda$ fixed (Right). The red circle highlights the highest accuracy obtained in each plot.}
\label{fig:sens}
\end{figure*}

FedDOSE achieves higher site-level and global accuracy than other \ac{FL} methods at nearly all sites across all three datasets. Some of the improvements were statistically significant based on two-sided Student's t-test and showed large effect sizes (Cohen’s d $\geq$ 0.8). The experimental results on ADHD-200, ABIDE-I, and ABIDE-II are summarized in Tables \ref{tab:adhd200}\footnote[1]{Results are reported as mean $\pm$ standard deviation. * indicates statistical significance (p-value \(<\) 0.05), \textasciicircum{} indicates a large effect size (\(d \ge 0.8\))}, \ref{tab:abide1}\footnotemark[1], and \ref{tab:abide2}\footnotemark[1], respectively. FedDOSE achieves the highest site accuracy at all sites except UCLA in ABIDE-I. Compared to FS\(^2\)G, the strongest baseline, FedDOSE improves site accuracy by about 4\% on ABIDE-II and 3\% on ADHD-200. On ABIDE-I, the two methods perform comparably. Since FS\(^2\)G does not explicitly model site heterogeneity, these gains suggest that FedDOSE's global alignment mechanism and decomposition of site differences are beneficial in multi-site settings. 

The Centralised baseline serves as a privacy-violating upper bound for the global FL model. In global accuracy, FedDOSE narrows this gap to about 2--3\% on ABIDE-I and at most 2\% on ADHD-200, nearly matching the Centralised model at site PKU. We observe similar trends in ABIDE-II, where FedDOSE comes within 1\% of the Centralised baseline at site EMC. FedDOSE's site accuracy exceeds the Centralised baseline at all sites, owing to local adaptation after federated training. Another consistent trend across all three datasets is that vanilla aggregation and simple prototype-based methods achieve the lowest performance. Brain network--modeled methods show modest improvements, while prototype-based personalization methods with alignment strategies and knowledge distillation achieve comparatively higher performance levels. These results suggest that the gains in FedDOSE's performance are likely systematic rather than dataset-specific.

\subsection{Ablation studies}

We ablated each component of FedDOSE on the ABIDE-I dataset. The results, summarized in Table \ref{tab:ablation}\footnotemark[1], show a substantial drop in performance when the phenotypic and scanner subspaces, along with their associated loss terms, are removed. This highlights the importance of decomposing site differences into learned subspaces. Otherwise, such confounded site factors may interfere with disease-related signals and degrade predictive performance. Furthermore, removing the reconstruction loss from the ModTKD block and applying a direct \ac{TKD} to the large DFC tensor led to a noticeable performance drop. Removing geometric alignment of local prototypes using Procrustes analysis and OT resulted in inferior performance compared to the full architecture. This confirms the effectiveness of our prototype alignment strategy in addressing site heterogeneity. Our ablation experiments demonstrate the importance of each component of FedDOSE for distinguishing \ac{ASD} participants from typically developing controls. 

\subsection{Sensitivity analysis}

\begin{figure*}[t]
\centering
\includegraphics[width=0.85\textwidth]{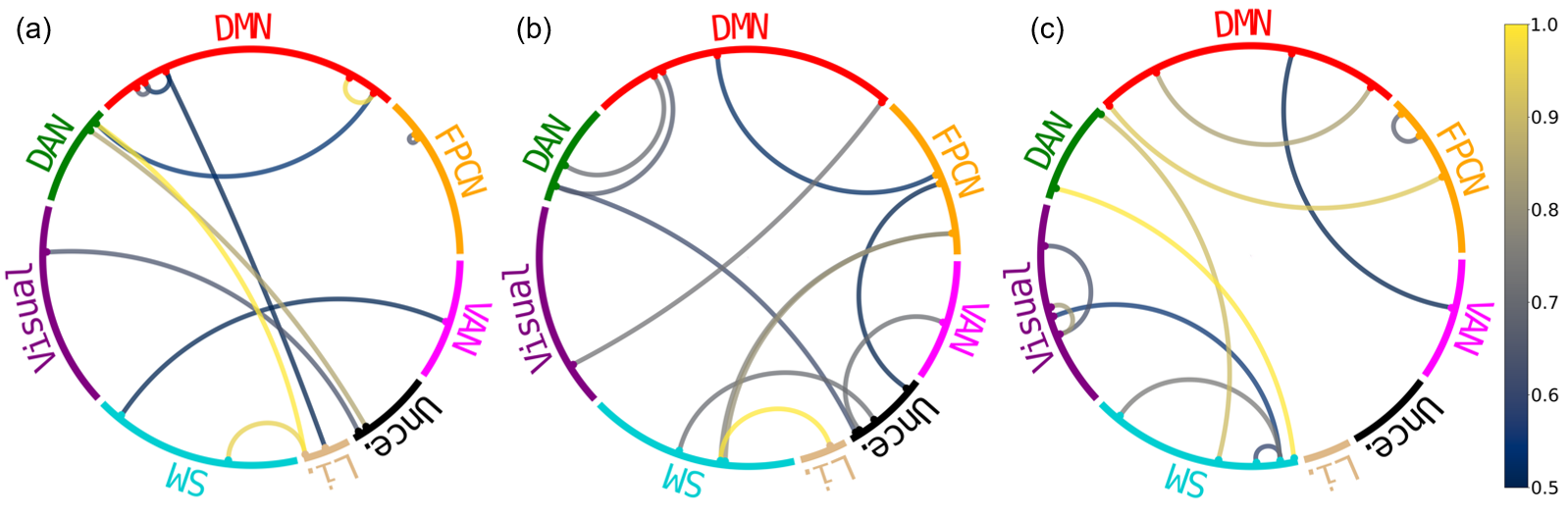}
\caption{Chord diagrams representing site-specific salient \ac{FC} connections of ABIDE-I. Each chord diagram shows only the 10 most salient connections for clarity. (a) NYU (b) UCLA (c) UM. Labels:- DMN: Default Mode Network; FPCN: Frontoparietal control network; VAN: Ventral Attention Network; Unce.: Uncertain; Li.: Limbic Network; SM: Sensorimotor Network; Visual: Visual network; DAN: Dorsal Attention Network}
\label{fig:chord}
\end{figure*}

We conducted a sensitivity analysis on the ABIDE-I dataset to evaluate the effects of varying the hyperparameters $\mu$, $\gamma$, and $\lambda$ within the FedDOSE framework. Evaluation was conducted by averaging the global accuracy across five seeds. We fixed one hyperparameter at its optimal value and varied the other two across multiple combinations. Our findings, shown in Figure \ref{fig:sens}, indicate that $\lambda$ and $\gamma$ are optimal at $0.2$ and $0.1$, respectively. Across all sites, performance with respect to $\lambda$ typically peaked at $0.2$ before declining. In contrast, global accuracy consistently decreased with increasing $\gamma$. $\mu$ showed an increasing trend up to $0.05$, after which it dropped for higher values. At site UM, $\mu$ was highest at $0.01$, which differed from the other two sites. Model performance generally degraded at extreme hyperparameter values. Based on this empirical evaluation, we set $\mu = 0.05$, $\gamma = 0.1$, and $\lambda = 0.2$ as the default values for downstream analyses.

\begin{table}[!ht]
\caption{linear probing three subspaces learned by FedDOSE on ABIDE-I. age is evaluated using r\(^2\), rest using accuracy}
\label{tab:probe}
\centering
\setlength{\tabcolsep}{2pt}
\resizebox{\columnwidth}{!}{
\begin{tabular}{lccccc}
\toprule
\textbf{Target Probe} & Disease & Gender & Age & Site & Scanner Type \\
\midrule
Disease (\(z_d\))
& \textbf{0.66\,\(\pm\)\,0.04} & 0.48\,\(\pm\)\,0.07 & 0.23\,\(\pm\)\,0.06 & 0.44\,\(\pm\)\,0.05 & 0.47\,\(\pm\)\,0.06 \\
Phenotype (\(z_b\))
& 0.42\,\(\pm\)\,0.06 & \textbf{0.59\,\(\pm\)\,0.06} & \textbf{0.56\,\(\pm\)\,0.11} & 0.49\,\(\pm\)\,0.04 & 0.45\,\(\pm\)\,0.05 \\
Scanner (\(z_v\))
& 0.39\,\(\pm\)\,0.07 & 0.49\,\(\pm\)\,0.06 & 0.19\,\(\pm\)\,0.08 & \textbf{0.58\,\(\pm\)\,0.04} & \textbf{0.60\,\(\pm\)\,0.03} \\
\bottomrule
\end{tabular}}
\end{table}

\begin{table*}[t]
\caption{Comparison of FedDOSE with baseline methods under cross-site validation on ABIDE-I dataset}
\label{tab:genacc}
\centering
\small
\setlength{\tabcolsep}{4pt}
\begin{tabular}{lcccccc}
\toprule
\textbf{Models} 
& \multicolumn{2}{c}{\textbf{NYU}} 
& \multicolumn{2}{c}{\textbf{UCLA}} 
& \multicolumn{2}{c}{\textbf{UM}} \\
\cmidrule(lr){2-3} \cmidrule(lr){4-5} \cmidrule(lr){6-7}
& \shortstack{Site Acc} & \shortstack{Genr. Acc}
& \shortstack{Site Acc} & \shortstack{Genr. Acc}
& \shortstack{Site Acc} & \shortstack{Genr. Acc} \\
\midrule
FedAvg \cite{mcmahan2017communication}
& 0.58 \(\pm\) 0.08*\textasciicircum & 0.49 \(\pm\) 0.09*\textasciicircum
& 0.65 \(\pm\) 0.08\,\;\textasciicircum & 0.55 \(\pm\) 0.07*\textasciicircum
& 0.62 \(\pm\) 0.08*\textasciicircum & 0.53 \(\pm\) 0.10*\textasciicircum \\
FedProto \cite{tan2022fedproto}
& 0.59 \(\pm\) 0.07*\textasciicircum & 0.52 \(\pm\) 0.08\,\;\textasciicircum
& 0.63 \(\pm\) 0.08*\textasciicircum & 0.57 \(\pm\) 0.09\,\;\textasciicircum
& 0.64 \(\pm\) 0.07*\textasciicircum & 0.56 \(\pm\) 0.09*\textasciicircum \\
TDPFed \cite{wang2022tensor}
& 0.60 \(\pm\) 0.06*\textasciicircum & 0.55 \(\pm\) 0.08\,\;\textasciicircum
& 0.66 \(\pm\) 0.06*\textasciicircum & 0.59 \(\pm\) 0.09\,\;\textasciicircum
& 0.67 \(\pm\) 0.07\,\;\textasciicircum & 0.61 \(\pm\) 0.09\,\;\textasciicircum \\
GAFD \cite{wang2025graph} 
& 0.62 \(\pm\) 0.06*\textasciicircum & 0.56 \(\pm\) 0.08\,\;\phantom{\textasciicircum}
& 0.69 \(\pm\) 0.08\,\;\textasciicircum & 0.59 \(\pm\) 0.06\,\;\textasciicircum
& 0.70 \(\pm\) 0.07\,\;\textasciicircum & 0.60 \(\pm\) 0.09\,\;\textasciicircum \\
FedGST \cite{mao2023fedgst}
& 0.65 \(\pm\) 0.06\,\;\textasciicircum & 0.58 \(\pm\) 0.09\,\;\phantom{\textasciicircum}
& 0.71 \(\pm\) 0.07\,\;\phantom{\textasciicircum} & 0.60 \(\pm\) 0.07\,\;\textasciicircum
& 0.71 \(\pm\) 0.06\,\;\textasciicircum & 0.62 \(\pm\) 0.08\,\;\textasciicircum \\
FedAli \cite{ek2024fedali}
& 0.66 \(\pm\) 0.05\,\;\textasciicircum & 0.59 \(\pm\) 0.07\,\;\phantom{\textasciicircum}
& 0.72 \(\pm\) 0.06\,\;\phantom{\textasciicircum} & 0.64 \(\pm\) 0.06\,\;\phantom{\textasciicircum}
& 0.73 \(\pm\) 0.06\,\;\phantom{\textasciicircum} & 0.66 \(\pm\) 0.08\,\;\phantom{\textasciicircum} \\
FedGMKD \cite{zhang2024fedgmkd}
& 0.68 \(\pm\) 0.06\,\;\phantom{\textasciicircum} & 0.60 \(\pm\) 0.06\,\;\phantom{\textasciicircum}
& 0.73 \(\pm\) 0.06\,\;\phantom{\textasciicircum} & 0.65 \(\pm\) 0.06\,\;\phantom{\textasciicircum}
& 0.73 \(\pm\) 0.07\,\;\phantom{\textasciicircum} & 0.66 \(\pm\) 0.06\,\;\phantom{\textasciicircum} \\
FS\(^2\)G \cite{mao2024toward}
& 0.69 \(\pm\) 0.05\,\;\phantom{\textasciicircum} & 0.61 \(\pm\) 0.05\,\;\phantom{\textasciicircum}
& \textbf{0.76 \(\pm\) 0.05\,\;\phantom{\textasciicircum}} & 0.66 \(\pm\) 0.06\,\;\phantom{\textasciicircum}
& 0.76 \(\pm\) 0.06\,\;\phantom{\textasciicircum} & 0.67 \(\pm\) 0.07\,\;\phantom{\textasciicircum} \\
FedDOSE 
& \textbf{0.71 \(\pm\) 0.06\,\;\phantom{\textasciicircum}} & \textbf{0.62 \(\pm\) 0.07\,\;\phantom{\textasciicircum}}
& 0.75 \(\pm\) 0.06\,\;\phantom{\textasciicircum} & \textbf{0.68 \(\pm\) 0.07\,\;\phantom{\textasciicircum}}
& \textbf{0.77 \(\pm\) 0.07\,\;\phantom{\textasciicircum}} & \textbf{0.69 \(\pm\) 0.07\,\;\phantom{\textasciicircum}} \\
\bottomrule
\end{tabular}
\end{table*}

\subsection{Evaluating subspace decoupling}

To demonstrate decoupling among disease, phenotype, and scanner subspaces, we performed linear probing experiments on each subspace. We pooled data from all sites in the ABIDE-I dataset and used a 60:40 train-test split. The experiments were repeated over five seeds and we report mean and standard deviation. We froze the trained model and extracted the three subspace embeddings. For each subspace, we fitted five independent linear probes on the training split. We used logistic regression for the categorical targets (disease, site, scanner type and gender) and linear regression for the continuous target (age). Each probe predicted a single target and was evaluated on the held-out test split. Repeating this for all three subspaces resulted in 15 probes as shown in Table \ref{tab:probe}. We observe that disease labels are predicted most accurately from the disease subspace. Likewise, gender and age are predicted most accurately from the phenotype subspace, while the scanner subspace achieves the highest accuracy in predicting site labels and site-specific scanner types. Overall, these results provide empirical evidence that the three subspaces learn distinct representations aligned with their intended objectives. 

We also performed three pairwise dependence tests using distance correlation (dCor) between the three subspace embeddings to assess their statistical dependence. The dCor between the disease and scanner embeddings was 0.224, whereas the dCor between the disease and phenotype embeddings was 0.295. The dCor between the phenotype and scanner embeddings was 0.361. The disease-related pairs exhibit lower distance correlation than the phenotype-scanner pair, suggesting weaker statistical dependence between the disease subspace and the other two subspaces. 

\begin{table*}[t]
\caption{Computational cost comparison on ABIDE-I dataset}
\label{tab:compcost}
\centering
\small
\setlength{\tabcolsep}{6pt}
\begin{tabular}{lccccc}
\toprule
\textbf{Metrics} & \textbf{FedDOSE} & \textbf{FedAvg} & \textbf{FedProto} & \textbf{FedGMKD} & \textbf{FS\(^2\)G} \\
\midrule
Communication Overhead & 2.3\% & 0\% (baseline) & 0.4\% & 2.1\% & 6.8\% \\
Wall Clock Time (s) & 48.8 \(\pm\) 6.3 & 31.2 \(\pm\) 4.7 & 33.3 \(\pm\) 5.6 & 40.1 \(\pm\) 5.9 & 43.4 \(\pm\) 5.2 \\
\bottomrule
\end{tabular}
\end{table*}

\subsection{Biomarker analysis}

Existing \ac{FL} studies in \ac{fMRI} analysis typically provide limited evaluation of framework interpretability and biomarker identification. To assess the effectiveness of FedDOSE, we performed a biomarker analysis at the modular level across five seeds and all participants in the ABIDE-I dataset. Saliency scores for FedDOSE were derived by computing the outer product of the first two columns of each module's spatial factor, which encodes its primary and secondary spatial patterns. Each chord diagram in Figure \ref{fig:chord} represents the top 10 salient connections. In Figure \ref{fig:chord}, connections involving the Sensorimotor Network (SM) and Dorsal Attention Network appear salient in sites NYU and UM, but are notably absent in UCLA. Similarly, connections between the SM and Limbic Network (Li.) appear salient in NYU and UCLA, but are absent in UM. This could suggest that these FC patterns may be age-specific, as UCLA generally recruits younger cohorts, whereas participants from UM tend to be older. These observations align with the existing literature on \ac{ASD} biomarkers \cite{zhou2024altered, wantzen2022eeg}. One factor that may support FedDOSE's identification of disease-related biomarker patterns is the orthogonality constraint in the \(\mathscr{L}_{\text{recon}}\) loss. By enforcing orthogonality among spatial patterns, it enables their cross-product interactions to capture diverse and salient relationships.

\subsection{Cross-site transfer analysis}

To assess the reliability of FedDOSE, we evaluated each site's locally adapted model on the test data from the other two sites after federated training and reported the averaged classification accuracy. We define this metric as generalization accuracy (Genr. Acc.) and used ABIDE-I dataset for this study. From Table \ref{tab:genacc}\footnotemark[1], we find that FedDOSE consistently achieved higher generalization accuracy than the baseline models. This demonstrates that FedDOSE can effectively identify and decompose site-specific signatures while capturing invariant \ac{ASD} features.

\subsection{Computational cost analysis}

To assess the practical overhead of FedDOSE, we compared its computational cost and training time with FedAvg \cite{mcmahan2017communication}, FedProto \cite{tan2022fedproto}, FedGMKD \cite{zhang2024fedgmkd}, and FS\(^2\)G \cite{mao2024toward} on the ABIDE-I dataset (Table \ref{tab:compcost}). Experiments were run over five seeds and we report mean \(\pm\) standard deviation. The training conditions were kept the same across all methods to ensure a fair comparison. Communication overhead refers to the amount of information exchanged between local sites and the server during each round, including model parameters and additional information such as class-conditional prototypes. Wall clock time is defined as the cumulative time taken by all sites to complete one round of federated training. Lightweight models such as FedAvg and FedProto have lower training times than more computationally intensive models such as FedDOSE, FedGMKD, and FS\(^2\)G. FedDOSE incurs a 2.3\% increase in per-round communication, calculated from the 5.8 MB of model parameters transmitted plus two class-conditional prototypes (one per class) totaling 0.136 MB per round. It requires 3.1 GB of GPU memory, making training feasible on standard consumer-grade GPUs. FedDOSE's per-round training time is about 12\% longer than FS\(^2\)G and about 56\% longer than FedAvg. This can be attributed to ModTKD at each site and the OT-based prototype alignment at the global level. However, this trade-off is offset by the observed performance improvements.

\section{Discussion and Conclusion}

This study presents several key findings. FedDOSE achieves competitive performance across two \ac{ASD} datasets and one \ac{ADHD} dataset compared with eight state-of-the-art \ac{FL} frameworks. It achieved statistically significant improvements with large effect sizes over earlier baselines such as FedAvg, FedProto, and TDPFed. Gains over stronger recent methods, including FedAli, FedGMKD, and FS\(^2\)G, were consistent but smaller. Our ablation results suggest that these gains stem primarily from three components. First, the supervised decomposition of site differences into disease, phenotype, and scanner subspaces prevents site-related artifacts from interfering with disease-relevant signals. Linear probing experiments further confirmed that each subspace predominantly encodes its intended factor. Second, combining \ac{TKD} with brain modularity preserves meaningful spatial network topology, while, unlike \cite{zhou2026cyclic}, the shared temporal factor captures how brain activity evolves over time and helps the modular core tensor learn robust representations. Third, aligning site prototypes via \ac{OT}-based Procrustes analysis before phenotype-weighted aggregation mitigates prototype heterogeneity across sites, which improved the performance of the global model.

Beyond classification accuracy, FedDOSE's locally adapted models transferred well to other participating sites in cross-site transfer experiments and maintained stable performance across the tested hyperparameter range, with modest degradation at extreme values. FedDOSE incurs a modest computational cost, adding only 2.3\% per-round communication overhead and requiring 3.1 GB of GPU memory, making it computationally feasible on a consumer-grade GPU. Our biomarker analysis further suggested that FedDOSE identifies site-specific connectivity patterns that are consistent with prior findings in the \ac{ASD} literature, providing preliminary evidence for the interpretability of our framework.

This study has some limitations. Compared with simpler \ac{FL} methods, FedDOSE introduces additional computational and communication overhead due to its tensor decomposition and prototype alignment steps. Since our focus was on methodological design rather than system deployment, we did not assess how the framework scales in a real-world federated network. FedDOSE requires phenotypic and scanner information to be available at every site and the biomarker analysis in this study was qualitative. Future work could focus on developing a client--server architecture, addressing deployment at scale and improving training efficiency, quantitatively validating the identified biomarkers, and extending the approach to other neurological disorders and imaging modalities beyond \ac{fMRI}.

\bibliographystyle{ieeetr}
\bibliography{reference}

\end{document}